\documentclass{amia}
\usepackage{graphicx}
\usepackage[labelfont=bf]{caption}
\usepackage[superscript,nomove]{cite}
\usepackage{color}
\usepackage{url}
\usepackage{amsmath}
\usepackage{comment}
\usepackage{booktabs}
\usepackage{threeparttable}
\usepackage{enumitem}

\usepackage{subfig}
\usepackage{booktabs}
\usepackage{multirow}
\usepackage{lipsum}

\begin{document}

\title{Generating (Factual?) Narrative Summaries of RCTs:\\
Experiments with Neural Multi-Document Summarization}

\author{Byron C. Wallace, PhD$^{1}$,
    Sayantan Saha, BS$^{1}$, \\
    Frank Soboczenski, PhD$^{2}$,
    Iain J. Marshall, MD, PhD$^{2}$,
    }

\institutes{
    $^1$Northeastern University, Boston, MA;
    $^2$King's College London, London}
\maketitle

{\bf Abstract} 

\textit{We consider the problem of automatically generating a narrative biomedical evidence summary from multiple trial reports. 
We evaluate modern neural models for abstractive summarization of relevant article abstracts from \emph{systematic reviews} previously conducted by members of the Cochrane collaboration, using the \emph{authors conclusions} section of the review abstract as our target.\footnote{We make this dataset available at: \url{https://github.com/bwallace/RCT-summarization-data}.}
We enlist medical professionals to evaluate generated summaries, and we find that summarization systems yield consistently fluent and relevant synopses, but these often contain factual inaccuracies.
We propose new approaches that capitalize on domain-specific models to inform summarization, e.g., by explicitly demarcating snippets of inputs that convey key findings, and emphasizing the reports of large and high-quality trials.
We find that these strategies modestly improve the factual accuracy of generated summaries.
Finally, we propose a new method for \emph{automatically} evaluating the factuality of generated narrative evidence syntheses using models that infer the directionality of reported findings.}


\section{Introduction}
\label{section:intro}

Biomedical \emph{systematic reviews} aim to synthesize all evidence relevant to a given clinical query \cite{sackett1997evidence,gough2017introduction}.
Such reviews typically comprise both quantitative and narrative summaries of the evidence.
The former is most often a statistical meta-analysis of the results reported in the constituent trials, which in turn informs the natural language interpretation provided in the latter. 
In Cochrane reviews,\footnote{\url{https://www.cochrane.org/}} brief narrative summaries communicating the main review findings are provided in structured abstracts in the \emph{Authors' conclusions} section. Below (left) is an example from a review of the evidence concerning the use of inhaled antibiotics for cystic fibrosis \cite{ryan2011inhaled}. We also provide the summary generated by one of the automated models we evaluate (right), given the abstracts of the included papers.  

\begin{center}
\begin{table}[h!]
\begin{tabular}{ p{.47\textwidth} p{.47\textwidth} } 
 \hline
 {\bf \emph{Authors' conclusions}} Inhaled antibiotic treatment probably improves lung function and reduces exacerbation rate, but a pooled estimate of the level of benefit is not possible. The best evidence is for inhaled tobramycin. More evidence, from trials of longer duration, is needed to determine whether this benefit is maintained and to determine the significance of development of antibiotic-resistant organisms. & {\bf \emph{Automatically generated summary}}
Inhaled antibiotics are effective in the treatment of Pseudomonas aeruginosa pulmonary infection in CF patients. However, there is a need for further randomised controlled trials to assess long-term safety and efficacy of inhaled antibiotics in patients with cystic fibrosis. Further trials are also needed to assess the effects of antibiotic treatment on morbidity and mortality. \\
 \hline

\end{tabular} 
\caption{Example \emph{Author conclusions} from a Cochrane systematic review abstract (left) and an automatically generated summary (right), conditioned on the set of clinical trial abstracts that informed the corresponding review. }
 \label{table:example}
\end{table}
\end{center}

Narrative summaries of clinical trials are invaluable for practitioners because they provide a concise, readable summary of all evidence relevant to the clinical question that motivated the corresponding review. These summaries are not verbatim distillations of the original trial reports, but can be considered as \emph{critical} summaries. The review authors should consider strengths and weaknesses of the source trials, see through any `spin' from the clinical trial authors, and emphasize the strongest evidence. The process of generating these summaries manually is lengthy and laborious. 

Consequently, summaries will not always be available for arbitrary clinical questions (even when relevant trial reports exist). 
Moreover, even where available they will often be out of date.
A system that could automatically summarize clinical trials literature would be capable of summarizing all evidence, on-demand.

In this work we evaluate state-of-the-art multi-document neural abstractive summarization models that aim to produce narrative summaries from the titles and abstracts of published reports of relevant randomized controlled trials (RCTs). 
We train these models using the \emph{Authors' conclusions} sections of Cochrane systematic review abstracts as targets, and the titles and abstracts from the corresponding reviews as inputs. 
We evaluate models both quantitatively and qualitatively, paying special attention to the \emph{factuality} of generated summaries. 
\subsection*{Related Work}
\label{section:related-work} 

\paragraph{Automatic Summarization and Question Answering for EBM}
This paper extends a thread of prior work on summarization for EBM \cite{demner2006answer,molla2010corpus,sarker2017automated}.
Demner-Fushman and Lin led a seminal effort on automatic question answering (QA) from the literature to aid EBM \cite{demner2006answer}.
This work on QA is adjacent to traditional summarization:
In their approach they aimed to extract snippets from individual articles relevant to a given question, rather than to \emph{generate} an abstractive summary of relevant abstracts, as is our aim here. 
Follow-up work on (extractive) QA over clinical literature has further demonstrated the promise of such systems \cite{cao2011askhermes}. 
For recent efforts in this vein, we point the reader to the latest BioASQ challenge iteration \cite{nentidis2019results}, which included a biomedical QA task. 
While related, we view the task of extractive biomedical QA as distinct from the more focussed aim of generating abstractive narrative summaries over relevant input abstracts to mimic narratives found in formal evidence syntheses (Table \ref{table:example}).

Directly relevant to this setting of biomedical systematic reviews, Molla \cite{molla2010corpus,molla2016corpus} introduced a dataset to facilitate work on summarization in EBM that comprises 456 questions and accompanying evidence-based answers sourced from the ``Clinical Inquiries'' section of the Journal of Family Practice. 
Sarkar \emph{et al.} \cite{sarker2017automated} surveyed automated summarization and EBM, respectively, highlighting the need for domain-specific multi-document summarization systems to aid EBM. 
In contrast to our approach, these prior efforts used comparatively small corpora, and pre-dated the current wave of the neural summarization techniques that have yielded considerable progress in language generation and (abstractive) summarization \cite{see2017get,lewis2019bart,zhang2019pegasus}.  

\paragraph{Neural Abstractive Summarization}
Automatic summarization is a major subfield in NLP\cite{maybury1999advances,nenkova2011automatic}. 
Much of the prior work on summarization of biomedical literature has used \emph{extractive} techniques, which directly copy from inputs to produce summaries.
However, narrative evidence synthesis is an inherently \emph{abstractive} task --- systems must generate, rather than simply copy, text --- as it entails communicating an overview of all available evidence.

Recent work on neural models has engendered rapid progress on abstractive summarization\cite{rush2015neural,lin2019abstractive}; we do not aim to survey this extensively here.
Illustrative of recent progress --- and most relevant to this work --- is the Bidirectional and Auto-Regressive Transformers (BART) model\cite{lewis2019bart}, which recently achieved state-of-the-art performance on abstractive summarization tasks.
Because it forms the basis of our approach, we elaborate on this model in Section \ref{section:methods}.

Despite progress in summary generation, evaluating abstractive summarization models remains challenging\cite{van2019best}. 
Automated metrics calculated with respect to reference summaries such as ROUGE\cite{lin2004rouge} provide, at best, a noisy assessment of text quality. 
Of particular interest in the setting of evidence syntheses is the \emph{factuality} of generated summaries: Here, as in many settings, users are likely to value accuracy more than other properties of generated text\cite{reiter:2020,reiter-belz-2009-investigation}. Unfortunately, neural models for abstractive summarization are prone to `hallucinations` that do not accurately reflect the source document(s), and automatic metrics like ROUGE may not capture this  \cite{maynez2020faithfulness}. 

This has motivated recent efforts to automatically evaluate factuality.
Wang \emph{et al.} proposed \emph{QAGS}, which uses automated question-answering to measure the consistency between reference and generated summaries \cite{wang2020asking}. 
Elsewhere, Xu \emph{et al.} \cite{xu2020fact} proposed evaluating text factuality independent of surface realization via Semantic Role Labeling (SRL). 
We extend this emerging line of work here by manually evaluating the factuality of summaries produced of clinical trial reports, and proposing a domain-specific method for automatically evaluating such narrative syntheses.

\section{Methods}

\subsection*{Data} \vspace{-.5em}
 We use 4,528 systematic reviews composed by members of the Cochrane collaboration (\url{https://www.cochrane.org/}). 
These are reviews of all trials relevant to a given clinical question. 
The systematic review abstracts together with the titles and abstracts of the clinical trials summarized by these reviews form our dataset.
All data was downloaded via PubMed (i.e., we use only abstracts).
The reviews include, on average, 10 trials each. The average abstract length of included trials is 245 words.
We use the ``authors' conclusions" subsection of the systematic review abstract as our target summary (75 words on average).
We split this data randomly into 3,619, 455, and 454 reviews corresponding to train, development (dev), and test sets, respectively.  The dataset is available at: \url{https://github.com/bwallace/RCT-summarization-data}.

\vspace{-.5em}
\subsection*{Models} \vspace{-.5em}
\label{section:methods}

We adopt Bidirectional and Auto-Regressive Transformers (BART) as our underlying model architecture\cite{lewis2019bart}. This is a generalization of the original BERT\cite{devlin2018bert} Transformer\cite{vaswani2017attention} model and pretraining regime in which self-supervision is not restricted to the objectives of (masked) token and next sentence prediction (as in BERT). Instead, BART is defined as an encoder-decoder model with an autoregressive decoder trained to `denoise' arbitrarily corrupted input texts. 
Masking tokens --- the original BERT objective --- is just one type of `corruption'. 
This permits use of additional corruption schemes (pretraining objectives), a property that we exploit in this work (Section \ref{section:additional-pretraining}). 
BART achieves strong performance on abstractive summarization tasks\cite{lewis2019bart}, which makes it particularly appropriate for our use here. 

BART defines a sequence-to-sequence network\cite{sutskever2014sequence} in which the \emph{encoder} is a bidirectional Transformer network and the \emph{decoder} is autoregressive (and hence amenable to language generation tasks such as summarization). 
One limitation of BART (and large neural encoder models generally) is that it imposes a limit on the number of input words that can be accepted due to memory constraints; for BART this limit is 1024. 
We discuss this further below.

We do not modify the BART architecture, but we explore new, domain-specific pretraining strategies and methods that entail modifying inputs. 
For the former, we propose and evaluate additional pretraining in which the objective is to construct abstracts of RCT articles from corresponding full-texts (Section \ref{section:additional-pretraining}). 
For the latter, we propose and evaluate a method in which we `decorate' input texts with annotations automatically produced by trained models, e.g., we explicitly demarcate (via special tokens) snippets in the input that seem describe interventions and key findings (Section \ref{section:inputs}). 
This is a simple (and as far as we aware, novel) means of incorporating prior information or instance meta-data in end-to-end neural summarization models.\footnote{Though the general idea of demarcating parts of inputs with special tokens for Transformers has been used for other tasks\cite{zhang2020paraphrase,rosset2020knowledge}.}

\vspace{-.5em}
\subsubsection*{Initialization and pre-training strategies}
\label{section:additional-pretraining}
\vspace{-.5em}

We use the BART-large version of BART,\footnote{Provided via the {\tt huggingface Transformers} library \cite{wolf2019huggingface}.} in which both the encoder and decoder are 12-layer Transformers. 
The `vanilla' variant of BART is initialized to weights learned under a set of denoising objectives that differ in how they corrupt the input text (which is then to be reconstructed). 
For example, objectives include token masking (as in the original BERT), `text infilling', and `sentence permutation' tasks \cite{lewis2019bart}. 
This pretraining is performed over a very large corpus comprising: {\tt BookCorpus}\cite{zhu2015aligning} and {\tt English Wikipedia}, {\tt CC-News}\cite{liu2019roberta}, {\tt OpenWebText},\footnote{\url{http://web.archive.org/ save/http://Skylion007.github.io/ OpenWebTextCorpus}} and {\tt Stories}\cite{trinh2018simple} (over 160GB of raw text in all). 
We verified via string matching that none of the target summaries (published Cochrane review abstracts) appeared in this pretraining corpora.
As a natural starting point for our task, we initialize BART-large weights to those learned via fine-tuning on the XSUM abstractive summarization corpus\cite{narayan2018don}.

With this as our starting point, we explored additional `in-domain` pretraining prior to learning to summarize trials.
Specifically we train BART to generate summaries from full-text articles. 
Specifically, we use $\sim$60k full-texts from the PubMed Central (PMC) Open-Access set that were classified as describing RCTs in humans by a previously developed model \cite{marshall2018machine}. 
Full-texts exceed the 1024 token budget imposed by BART, and so we alternate between selecting sentences from the start and end of the article text until we reach this limit.

\vspace{-.5em}
\subsubsection*{Modifying Inputs} \vspace{-.5em}
\label{section:inputs}

\begin{figure}
    \centering
    \includegraphics[scale=0.375]{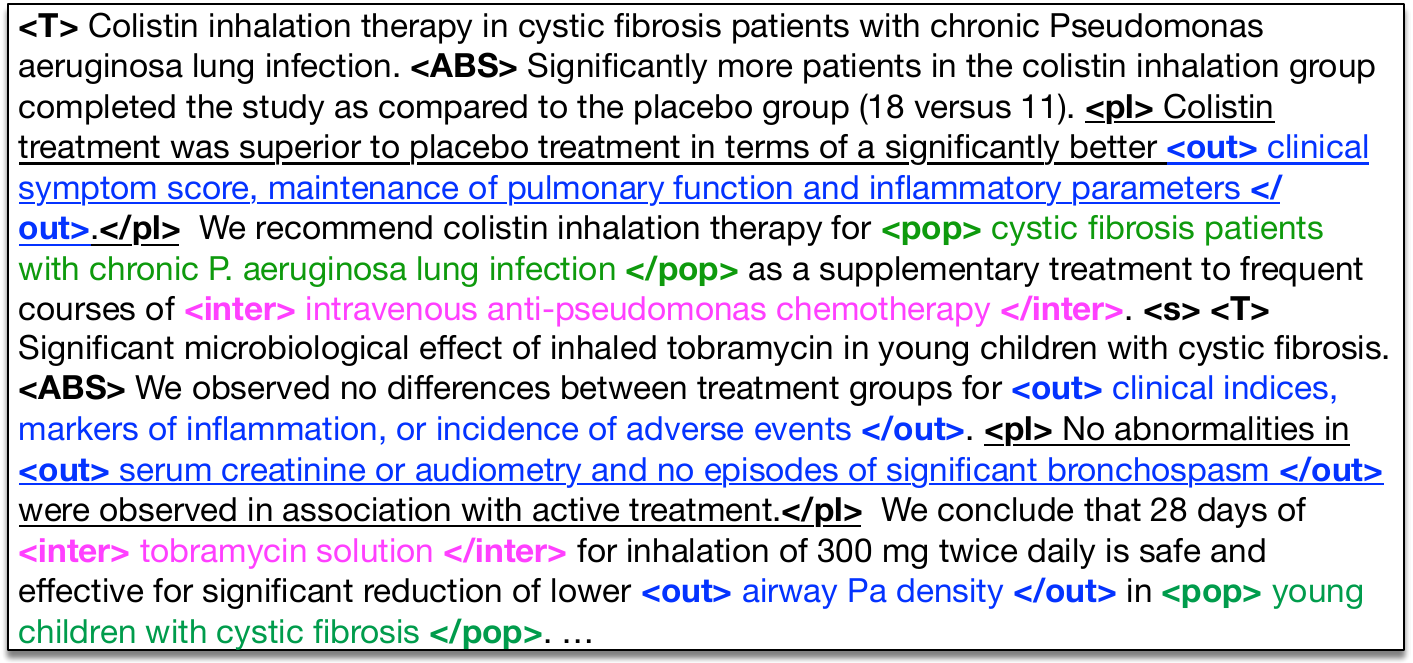}
    \caption{Input articles (here we show two for illustration) `decorated' using special tokens to demarcate automatically extracted salient attributes: \underline{$<$pl$>$} for `punchlines' sentences (those that seem to state the main finding), and snippets of text describing study populations $<$pop$>$, interventions $<$inter$>$, and outcomes $<$out$>$, respectively.}
    \label{fig:decoration-ex}
\end{figure}

Another important design choice concerns the inputs that we provide to the encoder component of BART. 
In the most straightforward use, we simply pass along subsets of the raw titles and abstracts. 
We demarcate titles, abstracts, and the start of new documents with with special tokens (`$<$T$>$', `$<$ABS$>$', `$<$S$>$'). 
Typically, only some of the abstracts associated with a given review will fit within the aforementioned token limit. 
We prioritize including titles, and then sentences from the beginnings and ends of abstracts. 
We select the latter in a `round-robin' (random) order from inputs, alternating between the starts and ends of abstracts, until the token budget is exhausted. 

\vspace{.2em}
\noindent {\bf Decoration} Prior work has investigated methods for automatically extracting key trial attributes from reports of RCTs, including descriptions of the study Populations, Interventions/Comparators, and Outcomes (the `PICO' elements) \cite{nye2018corpus} and identifying `punchline' snippets that communicate the main study findings \cite{lehman2019inferring}.
These key aspects of trials ought to figure prominently in summaries of the evidence. 
But in a standard end-to-end summarization approach, the model would have to implicitly learn to focus on these attributes, which seems inefficient. 

We propose a simple `decoration' strategy in which we explicitly demarcate snippets of text tagged by pretrained models as describing the aforementioned attributes. 
Decoration entails enclosing snippets (automatically) identified as describing the respective attributes within special tokens that denote these.
We provide an example (showing only two articles) in Figure \ref{fig:decoration-ex}. 
This preprocessing step is a simple mechanism to directly communicate to the encoder which bits of the text seem to provide information for the aspects of interest. 
To identify PICO elements, we use RobotReviewer\cite{marshall2016robotreviewer,marshall2017automating}. 
To identify punchline sentences, we fine-tuned a BioMed-RoBERTa model \cite{domains} on the Evidence Inference (2.0) dataset \cite{deyoung2020evidence}, which includes annotations of evidence-bearing (`punchline') sentences in RCT articles. 

\vspace{0.2em}
\noindent{\bf Sorting} Rather than treating all inputs equally, we might prefer to prioritize inclusion of evidence from large, high-quality studies. 
To operationalize this intuition, we consider a variant in which we greedily include tokens from abstracts ordered by sample size ($N$) scaled by an estimate of overall risk of bias (RoB) (a proxy for study quality). 
We infer both of these quantities automatically using RobotReviewer\cite{marshall2017automating,marshall2016robotreviewer}. 
Here RoB is the predicted probability of a study being at overall low risk of bias, based on the abstract text.

 \vspace{-.5em}
\subsection*{Design} 
 \vspace{-.25em}
We analyze the performance of five model variants that use elements of the above strategies (see Table \ref{table:main-results-dev-test-combined}). 
All are fine-tuned on the training set of Cochrane reviews. 
For `XSUM' we initialize BART to weights estimated on the XSUM abstractive summarization task \cite{narayan-etal-2018-dont}. 
For `Pretrain (PMC)' we continue pretraining over the PMC set as described above; all other models start from this checkpoint.
`Decorate' marks up the inputs as described above before passing them to the encoder (at train and test time).
`Sort by N$\cdot$RoB' greedily picks 1024 tokens by selecting for inclusion words from abstracts with the lowest (inferred) risk of bias, scaled by (extracted) sample size ($N$).

\vspace{0.25em}
\noindent{\bf Hyperparameters} During fine-tuning we used learning rate of $3$e$-5$. During decoding we used beam size of 4, a minimum target length of 65, we enabled early stopping, and we prevent three consecutive $n$-grams from repeating.
This largely follows the original BART paper\cite{lewis2019bart}; we did not systematically tune these hyperparameters.

\subsection*{Main outcome measurements} 
 \vspace{-.25em}
We measure summarization system performance using both automated and manual approaches.
For the former we use Recall-Oriented Understudy for Gisting Evaluation (ROUGE) \cite{lin2004rouge}, which relies on word overlaps between generated and reference summaries.
For the latter we enlisted medical doctors to annotate generated summaries with respect to relevance, plausibility (including fluency), and factuality (i.e., agreement with reference target summaries).
For this we built a custom interface; task instructions (with screenshots) are available here: \url{http://shorturl.at/csJPS}. 

As we later confirm empirically, ROUGE scores do not necessarily capture the factual accuracy of generated summaries, which is critical when generating evidence syntheses.
Manual evaluation of summaries can capture this, but is expensive, hindering rapid development of new models.
We propose a new approach to automatically evaluating generated narrative evidence syntheses with respect to the factuality of the findings they present. 
Specifically, we infer the reported directionality of the main finding in the generated and reference summaries, respectively, and evaluate the resultant level of (dis)agreement. 

To derive this automated metric we use the Evidence Inference dataset\cite{lehman2019inferring}, which comprises full-text RCT reports in which evidence-bearing snippets have been annotated, along with whether these report that the finding is a \emph{significant decrease}, \emph{significant increase}, or \emph{no significant difference} with respect to specific interventions, comparators, and outcomes. 
We simplify this by collapsing the first two categories, yielding a binary classification problem with categories \emph{significant difference} and \emph{no significant difference}.
Following DeYoung \emph{et al.}\cite{deyoung2020evidence}, we train a `pipeline' model in which one component is trained to identify `punchline' sentences within summaries, and a second is trained to infer the directionality of these findings. 
Both models are composed of a linear layer on top of BioMed-RoBERTa \cite{domains}.

Using these models we can infer whether reference and generated summaries appear to agree.
Specifically, we use the Jensen-Shannon Divergence (JSD) --- a measure of similarity between probability distributions --- between the predicted probabilities for \emph{sig. difference} and \emph{no sig. difference} from our inference model for the generated and reference summary texts, respectively. 
A low divergence should then suggest that the findings presented in these summaries is in agreement. We will call this measure \emph{findings-JSD}.

\section{Results}

\subsection*{Automated Evaluation} \vspace{-.5em}
\label{section:results-quant}

We report ROUGE-L scores with respect to the target (manually composed) Cochrane summaries, for both the development and test sets in Table \ref{table:main-results-dev-test-combined}. 
The methods perform about comparably with respect to this automatic metric. 
But ROUGE measures are based on (exact) $n$-gram overlap, and are insufficient for measuring the \emph{factuality} of generated texts\cite{maynez2020faithfulness,kryscinski2019neural}.
Indeed, we find that the summaries generated by all variants considered enjoy strong fluency, but the key question for this application is whether generated summaries are \emph{factually correct}. 
Below we confirm via manual evaluation that despite achieving comparable ROUGE scores, these systems vary significantly with respect to the factuality of the summaries that they produce.

\begin{table*}
    \centering
    \footnotesize
    \begin{tabular}{llllrr}
        Name & Initialization   &  (Additional) Pretraining &            System inputs & \multicolumn{1}{c}{ROUGE-L ({\tt dev})} & \multicolumn{1}{c}{ROUGE-L ({\tt test})} \\
        \midrule
        XSUM & \emph{XSUM} & None & Titles and abstracts & 0.264 & 0.265 \\
        Pretrain (PMC) & \emph{XSUM} & PMC RCTs & Titles and abstracts & 0.263 & 0.269 \\
        Decorate & \emph{XSUM} & PMC RCTs & Decorated  & 0.268 & 0.266 \\
        Sort by N$\cdot$RoB & \emph{XSUM} & PMC RCTs & Sorted by $N \cdot$ RoB & 0.267 & 0.267 \\
        Decorate and sort & \emph{XSUM} & PMC RCTs & Decorated and sorted & 0.265  &  0.265\\
        \bottomrule
    \end{tabular} \vspace{-.5em}
    \caption{Model variants and ROUGE-L measures over the {\tt dev} and {\tt test} sets. (Results for ROUGE-1 and ROUGE-2 are qualitatively similar.) `PMC RCTs' is shorthand for our proposed strategy of (continued) pretraining to generate abstracts from full-texts for all RCT reports in PMC. All model variants aside from `XSUM' start from the Pretrain (PMC) checkpoint.}
    \label{table:main-results-dev-test-combined}
\end{table*}

\subsection*{Manual Evaluation} \vspace{-.5em}
\label{section:results-manual}

Manual annotation was performed for 100 reference Cochrane reviews and the 5 systems described above.
Annotators were shown summaries generated by these systems in turn, \emph{in random order}.
Randomization was performed independently for each review (i.e., for each reference summary).
Annotators did not know which system produced which summaries during assessment.
We asked four questions across two pages about generated summaries.

The first page displayed only the generated summary, and asked annotators to appraise its \emph{relevance} to the topic (the title of the corresponding systematic review) on a 3-point ordinal scale ranging from mostly off-topic to strongly on-topic. 
The second question on the first page concerned `semantic plausibility', intended to measure whether the generated text is understandable, coherent, and free from self-contradictory statements. 
This assessment was on a five-point (Likert) scale. 

Following these initial evaluations, annotators were asked two additional questions to assess the factuality of generated summaries with respect to the reference. 
The first concerned the direction of the reported finding in the reference summary (e.g., did the authors conclude the intervention being investigated beneficial compared with the control?).
The second question then asked the annotator to assess the degree to which the generated summary agreed with the reference summary in terms of these key conclusions. 
Both of these judgments were collected on Likert-scales. 

\begin{table*}
    \centering
    \footnotesize
\begin{tabular}{lrrr}
\toprule
System variant &  Relevance $>$2 &  Fluency $>$3 &  Factuality $>$3 \\
\midrule
 XSUM &            96 &          90 &             40 \\
  Pretrain (PMC) &            98 &          97 &             34 \\
  Decorate &            98 &          96 &             54 \\
 Sort by N $\cdot$ RoB &            96 &          93 &             46 \\
Decorate and sort &            93 &          88 &             47 \\
\bottomrule
\end{tabular}
\caption{Counts of generated summaries out of 100 assessed by MD 1 as exhibiting high relevance (3/3); good to very good fluency ($>$3/5); and moderate to strong factual agreement with reference summaries ($>$3/5).}
\vspace{-.5em}
\label{table:manual-results-counts}
\end{table*}

One author (IJM; a medical doctor with extensive experience in evidence-based medicine) performed the above evaluation on a `pilot' set of 10 reference reviews, yielding 50 system judgements in all. He did not know which systems produced which outputs. 
This pilot set was used to assess agreement with additional prospective annotators, who we recruited via the Upwork platform\footnote{\url{http://www.upwork.com}}.

We hired three candidate annotators (all with medical training) to assess the system summaries for pilot set of reviews appraised by IJM. 
Only one of these candidates provided reliable annotations, as determined by agreement with the reference set of assessments.\footnote{We assessed this both quantitatively and qualitatively. Rejected candidates provided uniformly high scores for all generated summaries, even in cases where, upon inspection, these were blatantly in disagreement with the reference summary.} 
Scores provided by the successful annotator (who we will refer to as `MD 1') achieved 0.535 linearly weighted $\kappa$ with reference annotations concerning `factuality', the hardest task, indicating reasonable agreement. 
IJM also subsequently evaluated all cases in this set where the label he had provided disagreed with MD 1's assessment (still blinded to which systems produced the corresponding summaries). 
These were determined reasonable disagreements, given that the task is somewhat subjective as currently framed.

\begin{figure}%
    \centering
    \subfloat[Scores from the MD hired via Upwork (MD 1) over 100 unique reference summaries (500 summary annotations).]{{\includegraphics[width=0.4\linewidth]{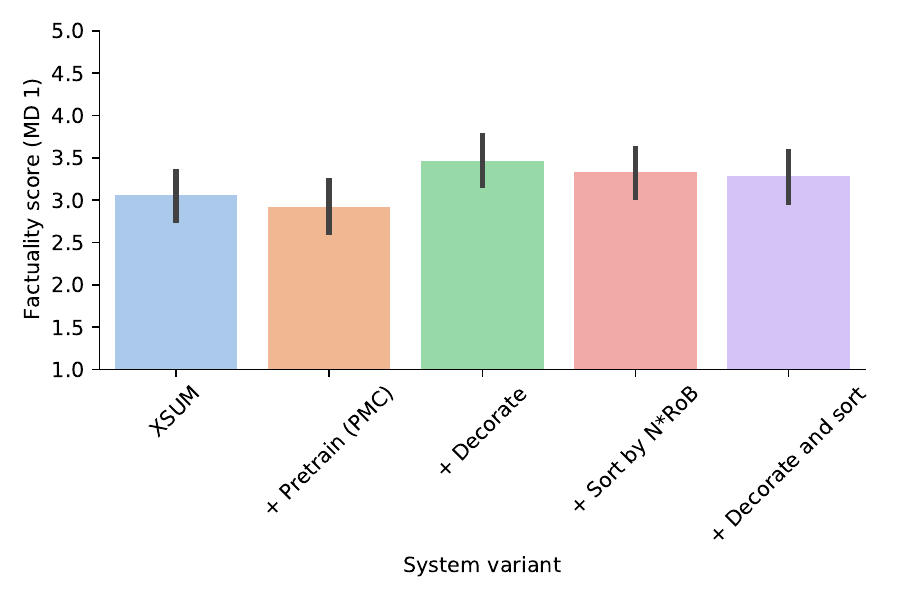} }}%
    \qquad
    \subfloat[Scores from co-author (and MD) IJM over a subset of 20 unique reviews (100 summary annotations).]{{\includegraphics[width=0.4\linewidth]{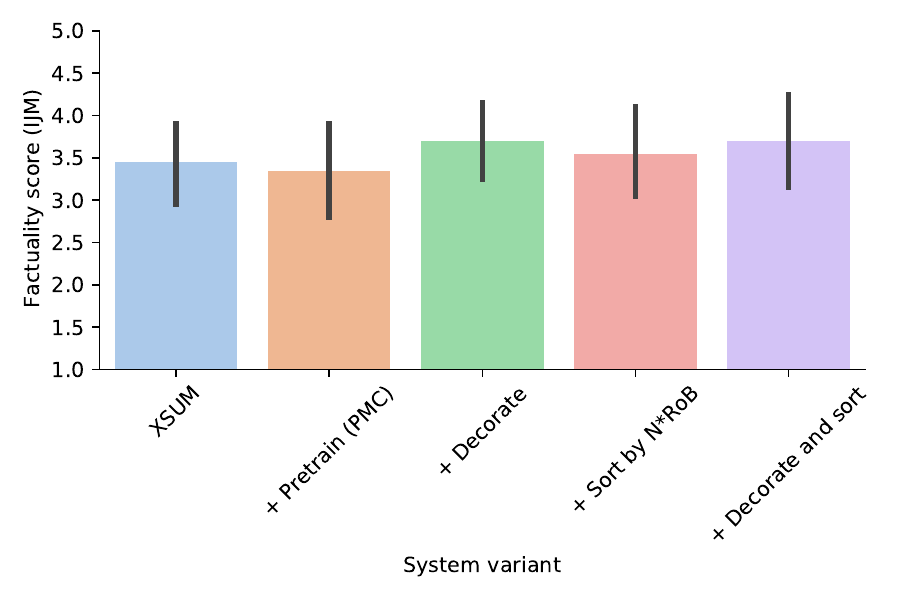} }}%
    \vspace{-.5em}
    \caption{Factuality assessments performed by an individual with medical training for five systems over 100 unique reference summaries from the {\tt dev} set (a), and by co-author and MD IJM over a small subset of twenty of these (b). All strategies except `XSUM' start from the model checkpoint after PMC pretraining. We first evaluate the `decoration' and sorting strategies (Section \ref{section:inputs}) independently, and then in combination; system names are consistent with Table \ref{table:main-results-dev-test-combined}.}%
    \label{fig:fact_scores}
    \vspace{-.25em}
\end{figure}

In total we collected assessments across the five system variants considered with respect to 100 unique corresponding Cochrane reference summaries from MD 1; for this we paid about \$1,500 USD. 
As a second (post-hoc) quality check, IJM evaluated an additional randomly selected subset comprising 10 reference reviews.
Agreement concerning relevance (80\% exact agreement) and fluency (68\% accuracy) remained high, as in the pilot round.
However, in contrast to the pilot set, agreement concerning factuality on this subset was ostensibly poor (linearly weighted $\kappa$=0.04); on average IJM scored systems higher in factuality by 1.62 points. 
As above, IJM therefore evaluated all instances for which disagreement was $\geq 2$ (again keeping blinding intact). 
This process again revealed predominantly reasonable subjective differences on this small set in assessing the level of agreement between the findings communicated in the generated and reference summaries, respectively.
MD 1 consistently rated lower factuality scores than IJM --- assigning lower numbers across the board --- but relative rankings seem to broadly agree (Figure \ref{fig:fact_scores}).

This disagreement suggests that in future work we should work to improve the annotation task framing and guidance.
The most common disagreement occurred in cases where the reference summary described a lack of reliable evidence on a topic, but \emph{hinted} cautiously that there was some small, or low quality evidence in favor of an intervention. If an automated summary only described a lack of reliable evidence on the topic, it was ambiguous whether the overall poor state of evidence should be scored (in this instance, showing perfect agreement), or by how much the automated summary should be penalized for missing the tentative conclusion of possible benefit.

Nonetheless, in light of strong agreement on \emph{other} rated aspects and our manual assessments of all substantial disagreements, we feel reasonably confident that MD 1 provided meaningful scores, despite the low quantitative measure of agreement on the second randomly selected set. 
And regardless, the broad trends across systems agree when the annotations from the two annotators are analyzed independently (Figure \ref{fig:fact_scores}).

All systems received consistently high relevance scores from MD 1 (mean scores for system summaries produced by different systems over the 100 reviews range from 2.73 to 2.79, out of 3), and `semantic plausibility' scores (range: 4.47 to 4.64 across systems, out of 5).
Table \ref{table:manual-results-counts} reports the counts of `good quality' summaries with respect to the aforementioned aspects, as judged by MD 1. 
We can see that systems struggle to produce factual summaries.

Figure \ref{fig:fact_scores} (a) reports the mean factuality scores provided by MD 1 for the respective model variants. 
The proposed `decorating' strategy yields a statistically significant improvement over the baseline PMC pretraining strategy (2.92 vs 3.46; $p\approx0.001$ under a paired $t$-test). 
Note that this is the appropriate comparison because the `+ Decorate' model starts from the PMC pretrained checkpoint.
Sorting inputs such that the encoder prioritizes including abstracts that describe large, high-quality studies (given the 1024 token budget imposed by BART) also increases factuality, albeit less so (2.92 vs 3.33; $p \approx 0.01$). 
Figure \ref{fig:fact_scores} (b) presents the factuality scores provided by IJM over a small subset of the data (20 unique reviews in all). 
The pattern is consistent with what we observed in MD 1's scores in that `decoration' yields increased factuality (mean score of 3.35 vs 3.70).\footnote{Though given the small sample of 20 reviews that IJM annotated neither difference is statistical significant when considering only these labels.}

 \vspace{-.5em}
\subsection*{Automatically Assessing the Factuality of Evidence Synopses} \vspace{-.5em}
\label{section:results-auto}

ROUGE scores do not vary much across model variants, but this probably mostly reflects the fluency of summaries --- which was also manually assessed to be uniformly good across systems. 
ROUGE (which is based on word overlap statistics) does not, however, seem to capture \emph{factuality}, which is naturally of central importance for evidence synthesis.
We tested this formally using annotations from MD 1: We regressed factuality judgements (ranging 1-5) on ROUGE-L scores (including an intercept term), over all annotated summaries. 
The trend is as we might expect: larger ROUGE-L scores are associated with better factuality ratings, but the correlation is not significant ($p \approx 0.18$).

We are therefore reliant on manual factuality assessments as we work to improve models.
Performing such evaluations is expensive and time-consuming: Collecting annotations over 100 instances for this work cost nearly \$2,000 USD (including payments to collect `pilot' round annotations) and investing considerable time in documentation and training.
Relying on manual assessments will therefore substantially slow progress on summarization models for evidence synthesis, motivating a need for automated factuality evaluation such as the \emph{findings-JSD} measure proposed above.

Repeating the regression we performed for ROUGE-L, we can measure whether findings-JSD correlates with manual factuality assessments. 
We define a regression in which we predict factuality scores on the basis of the JSD scores.
We find a statistically significant correlation between these with an estimated coefficient of -1.30 (95\% CI: -1.79 to -0.81; $p < 0.01$), implying that the larger the disagreement concerning whether the summaries report a significant effect or not (measured using JSD), the lower the factuality score, as we might expect.

This result is promising. But despite the significant correlation this automated metric has with manual assessments, it is not strong enough to pick up on the differences between strategies.
In particular, repeating the $t$-test on findings-JSD scores for the pretaining and decorating strategies yields a $p$-value of $0.40$, i.e., the measure fails to meaningfully distinguish the latter from the former with respect to factuality. 
We conjecture that this is because while the measure significantly correlates with human assessments, it does so only modestly ($R^2=0.05$).
We therefore conclude that this strategy constitutes a promising avenue for automatically assessing the factuality of generated summaries, but additional work is needed to define a measure that enjoys a stronger correlation with manual assessments.

\vspace{-0.35em}
\section{Discussion}  \vspace{-0.35em}

Above we proposed variants of modern neural summarizaiton models in which we: Perform additional in-domain pretraining (over the RCTs in PMC); `decorate' inputs with automatically extracted information (e.g., population descriptions and evidence-bearing sentences); and sort inputs to prioritize passing along large and high-quality trials (given the limit on the length of the model input imposed by the transformer model we use). 

We evaluated these models across key aspects, including relevance, `semantic plausibility', and factuality. 
All systems we considered yielded highly fluent and relevant summaries. 
But manual analysis of generated and corresponding reference summaries revealed that the \emph{factuality} of these systems remains an issue.
The proposed decoration and sorting strategies both yielded modest but statistically significant improvements in assessed factuality.

Annotators exhibited disagreement when evaluating factuality.
We believe this in part reflects the inherent difficulty of the task, but in future work we hope to improve the annotation protocol to reduce subjectivity and improve agreement.
For example, being more explicit in the levels of disagreement that should map onto specific numerical scores and providing more detailed instructions regarding this may improve inter-rater agreement, as might explicitly differentiating between the factuality of \emph{strength} of evidence and the reported directionality of the finding. 

ROUGE scores --- commonly used to automatically evaluate summarization systems --- did not significantly correlate with factuality assessments here.
We proposed a method for automatically evaluating the factuality of narrative evidence syntheses, findings-JSD, using models to infer the reported directionality of findings in generated and reference summaries.
This measure significantly (though weakly) correlates with manual assessments of factuality.
We view this as a promising direction to pursue going forward to facilitate automatic evaluation of evidence synopses, which in turn would support continued development of automated summarization systems for evidence synthesis.

\section{Conclusions} \vspace{-0.45em}

We have demonstrated that modern neural abstractive summarization systems can generate relevant and fluent narrative summaries of RCT evidence, but struggle to produce summaries that accurately reflect the underlying evidence, i.e., that are \emph{factual}.
We proposed new approaches that modestly improve the factuality of system outputs, and described a metric that attempts (with some success) to automatically measure factuality, suggesting directions for future work.
The multi-document summarization dataset is available: \url{https://github.com/bwallace/RCT-summarization-data}. 

\vspace{-0.25em}
\section*{Acknowledgements}  
\vspace{-.45em}
This work is funded by the National Institutes of Health (NIH) under the National Library of Medicine, grant R01-LM012086.
We thank Ani Nenkova for helpful comments concerning evaluation.

\setlength\itemsep{-0.1em}

\bibliographystyle{vancouver}
\bibliography{main}
\end{document}